\PassOptionsToPackage{svgnames}{xcolor} 
\documentclass{article} % For LaTeX2e
\usepackage[final]{colm2025_conference}
\usepackage{microtype}
\usepackage{hyperref}
\usepackage{url}
\usepackage{booktabs}

\usepackage{lineno}
\usepackage{adjustbox}
\usepackage{xspace}

\usepackage{amsmath}
\usepackage{amssymb}
\usepackage{mathtools}
\usepackage{amsthm}
\usepackage{enumitem}
\usepackage{soul}
\usepackage{graphicx}
\usepackage{wrapfig}

\newcommand{\name}{OverFill\xspace}

\title{OverFill: Two-Stage Models for Efficient Language Model Decoding}

\author{Woojeong Kim, Junxiong Wang, Jing Nathan Yan, Mohamed Abdelfattah, \\ 
\textbf{Alexander M. Rush} \\[0.4em]
Cornell University \\
\texttt{\{wk247, jw2544, jy858, Mohamed, arush\}@cornell.edu} \\
}

\begin{document}

\ifcolmsubmission
\linenumbers
\fi

\maketitle

% sections
\begin{abstract}
Large language models (LLMs) excel across diverse tasks but face significant deployment challenges due to high inference costs. LLM inference comprises \textit{prefill} (compute-bound) and \textit{decode} (memory-bound) stages, with decode dominating latency particularly for long sequences. Current decoder-only models handle both stages uniformly, despite their distinct computational profiles. We propose \name, which decouples these stages to optimize accuracy-efficiency tradeoffs. \name begins with a full model for prefill, processing system and user inputs in parallel. It then switches to a dense pruned model, while generating tokens sequentially. Leveraging more compute during prefill, \name improves generation quality with minimal latency overhead. Our 3B-to-1B \name configuration outperforms 1B pruned models by 83.2\%, while the 8B-to-3B configuration improves over 3B pruned models by 79.2\% on average across standard benchmarks.
\name matches the performance of same-sized models trained from scratch, while using significantly less training data. Our code is available at \url{https://github.com/friendshipkim/overfill}.
\end{abstract}
\section{Introduction}
\label{introduction}
Large language models (LLMs) have achieved remarkable success on a broad spectrum of tasks, from question answering to code generation. Yet, their massive parameter counts pose significant challenges for practical deployment, with \emph{inference} emerging as a chief bottleneck. In modern LLMs, inference typically comprises two stages: \textit{prefill} and \textit{decode}. The prefill stage, where all input tokens are processed in parallel to build a Key-Value (KV) cache, is usually \emph{compute-bound}: the performance is primarily limited by the utilization of computational units. The subsequent decode stage is \emph{memory-bound}, where it generates each output token autoregressively. The main bottleneck here is repeatedly loading the model's large feed-forward (FFN) layers into memory.

However, current LLM architectures do not exploit the distinct computational characteristics of these two stages. 
Our motivation stems from the distinct computational profiles of the prefill and decode stages, which have led to a growing trend of disaggregating them. The first line of work~\citep{zhong2024distserve, patel2024splitwise} is system-oriented, focusing on stage-specific resource allocation and parallelism while still using a single model across both stages. 
% Prior system-oriented methods~\citep{zhong2024distserve, patel2024splitwise} streamline the pipeline of prefill and decode but still use the full model for both stages, thereby incurring high memory overhead for each generated token.
The second line of work (algorithmic-oriented)~\citep{nair2024tandem, bergner2024think} explores using models of different sizes for each stage, but often requires complex frameworks or delivers only small accuracy improvements. The question of interest is how to \emph{decouple} the prefill and decode stages to achieve a stronger balance of accuracy and efficiency.

\begin{figure}[!t]
    \centering    \includegraphics[width=0.76\columnwidth]{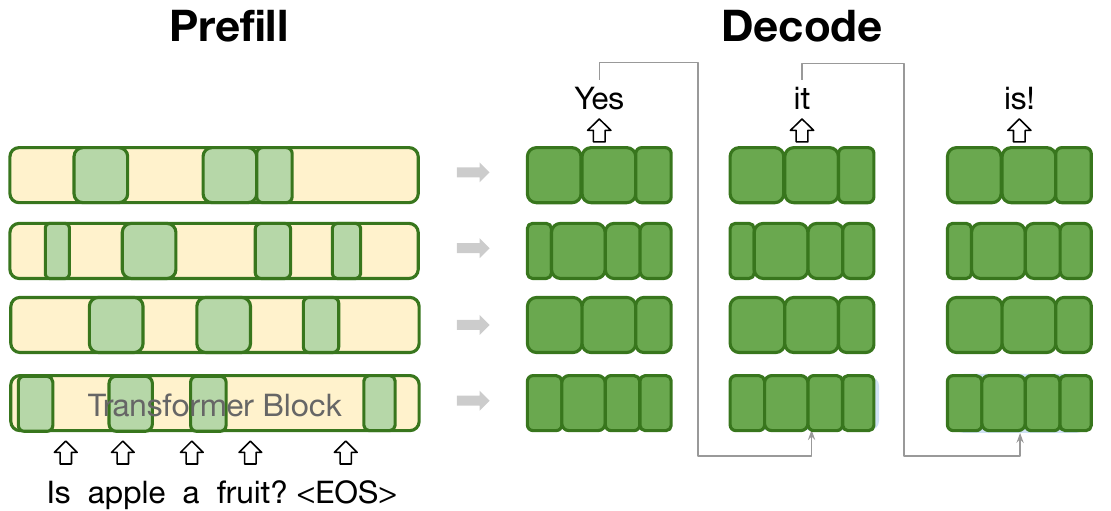}
    \caption{Overview of \name. \name uses the full model for prefill and a pruned model for sequential decoding. The yellow blocks represent the full model used for prefill. The decoder is initialized by selecting important channels (green blocks) from the full model. The blocks on the decoder side are in darker shades because they are updated during \name training. However, the full prefill model is kept frozen.}
    \label{fig:prefillvsdecode}
    % \vspace{-1em}
% \vskip -0.2in
\end{figure}

We propose \textbf{OverFill}, a two-stage system that dedicates maximal capacity to the prefill stage, while pruning the costly decode stage to reduce parameter loading. Specifically, the larger model is used only once to process the user input into a vector representation, and a smaller, pruned model subsequently handles token-by-token generation. As a result, \name drastically cuts the memory footprint and latency of decoding, especially for longer sequences. While it incurs a small prefill overhead compared to the standalone pruned model, this difference in cost is negligible compared to the dominant decoding latency.
Moreover, \name starts with a single model and prunes it to build a smaller decoder, eliminating the extensive process of aligning two distinct models. Since only the small decoder is updated during training, \name is compatible with the original full model during serving.
% Moreover, \name employs a single set of weights across both stages, offering the flexibility to use any single language model and eliminating the extensive process of aligning two distinct models. 
Importantly, \name is end-to-end trainable and does not require additional modules, allowing it to be optimized like any standard transformer architecture. Figure~\ref{fig:prefillvsdecode} illustrates the overview of our framework.\

% \wj{we need to be careful because we improve the accuracy in pruned scale, but can't match full} 
% By leveraging a full model for the initial context, \name preserves high accuracy on tasks such as QA, math, and chain-of-thought reasoning. Meanwhile, pruned decoding avoids the repeated overhead of loading large parameters. 

% The core idea behind OverFill is to exploit the computational asymmetry between prefill and decode. During prefill, the model processes the ``hard" information embedded in the user input, such as the critical elements of a question in a QA task. Using a stronger model for prefill allows for better encoding of this information, which in turn improves the overall generation quality.

% In terms of efficiency, OverFill achieves significant gains compared to full-model decoding in terms of memory usage and latency. It incurs a small prefill overhead compared to the standalone pruned model, yet this cost is negligible compared to the dominant decoding latency. This is more prominent for longer sequences, scenarios increasingly common with chain-of-thought reasoning. Additionally, as recent studies emphasize generating multiple candidates or exploring more decoding paths, the efficiency of small decoders becomes even more advantageous.

We validate \name in diverse decoder scales and compare \name accuracy on tasks such as question answering, math, and chain-of-thought reasoning. 
In a 3B-to-1B configuration, \name outperforms standalone pruned models by 83.2\% and instruction-tuned model by 52.6\% in average. \name matches or even outperforms similarly sized models trained from scratch, while using significantly fewer training tokens.
We also demonstrate \name is pareto-optimal in both scales. \name achieves significant accuracy gain over standalone small models while posing minimal latency overhead. These efficiency gains become even more pronounced when long outputs or multiple candidates are generated, as the pruned model remains active throughout the autoregressive process.

\section{Related work}
\label{related_work}

\paragraph{Model compression.}

Pruning enhances the efficiency of LLMs by removing model components that contribute the least to the output. Structured pruning eliminates entire groups of parameters, such as channels, attention heads~\citep{dery2024everybody, ma2023llm, xia2023sheared, ashkboos2024slicegpt} or layers~\citep{men2024shortgpt, yang2024laco, kim2024shortened}. Structured pruning results in a more compact model while preserving the underlying architecture, keeping it hardware-friendly. 
In this work, we adopt width pruning~\citep{dery2024everybody, ma2023llm, xia2023sheared, ashkboos2024slicegpt}, which preserves accuracy better than depth pruning. We specifically use the approach from \citet{sreenivas2024llm} but without the additional KL loss term, focusing on optimizing pruning to balance performance and computational efficiency.

% Unlike unstructured pruning, which removes individual weights in a sparse and often hardware-unfriendly manner,

% A main focus of pruning in recent works is in depth pruning~\citep{men2024shortgpt, yang2024laco, kim2024shortened} assess the importance of individual transformer blocks, pruning or skipping those less critical. Typically, the initial and final layers of the model are considered crucial, while intermediate layers are candidates for pruning. However, this approach introduces layer mismatches with the original model, potentially complicating model adaptation and finetuning.

% We adopt width pruning which is shown to preserve accuracy better than depth pruning.
% \citet{dery2024everybody, ma2023llm, xia2023sheared, ashkboos2024slicegpt} introduce saliency metrics and pruning strategies targeting width dimensions, including embedding channels, attention heads, and intermediate MLP channels. These approaches often rely on learning width masks using second-order derivatives, which can be computationally and memory intensive.

% In this work, we adopt the pruning approach proposed by \citet{sreenivas2024llm}, focusing on width dimensions at LLM scales. Unlike \citet{sreenivas2024llm}, we do not incorporate an additional KL loss term from the teacher (full) model during adaptation. Instead, we focus on optimizing the pruning process to maintain performance while minimizing computational overhead.

Quantization~\citep{frantar2022gptq, lin2024awq, xiao2023smoothquant} is another effective model compression method. Standard quantization methods accelerate both prefill and decode. In our two-stage decoding process, we can also consider high-precision prefill and low-precision decode, which we leave for future work.

% However, this is an orthogonal improvement that could further enhance our performance, which we leave as future work.

\paragraph{Decoding targeted speedups.}
Various methods tackle the serial bottleneck in LLM decoding. Speculative decoding~\citep{leviathan2023fast} leverages available compute to propose tokens in parallel using a small draft model. Researchers have explored specialized draft models~\citep{sun2021instantaneous, xia2023speculative} and subnetworks of the target model~\citep{schuster2022confident, elhoushi2024layerskip, zhang2023draft, liu2024kangaroo, ankner2024hydra}. Among these, \citet{du2024glide, li2024eagle} are particularly relevant to our work as they reuse target model representations to enhance drafting. Unlike speculative decoding, our approach eliminates rollbacks and calls the large model only once during prefill, avoiding parallel execution with the small model during decoding. This significantly reduces memory usage. We provide a theoretical analysis in the Appendix.

\paragraph{KV cache compression.}
Many studies have explored KV cache compression to address memory bottlenecks with heavy batching and long contexts, using methods like token eviction~\citep{xiao2023efficient, zhang2023h2o, adnan2024keyformer}, quantization~\citep{sheng2023flexgen, liu2024kivi}, and prompt compression~\citep{pan2024llmlingua, wingate2022prompt}. 
Our approach targets scenarios where loading weights is the primary memory bottleneck.
While KV cache can dominate memory usage in certain scenarios, this typically occurs only at very long sequence lengths when using smaller models. For instance, in a 7B parameter model with batch size 4, model weights is the primary bottleneck up to 5K tokens~\citep{adnan2024keyformer}.
Several works target the opposite challenge as ours: reducing prefill costs for very long contexts, where prefill becomes costly. These methods include architectural modifications~\citep{sun2024you}, chunking~\citep{zeng2024memorize}, token dropping~\citep{fu2024lazyllm}, and prompt packing~\citep{zhao2024prepacking}. Notably, such KV cache compression and prefill acceleration approaches can be applied on top of our method for further optimization.

\section{Method}
% \sasha{let's emphasize the simplicity of our system, only mention things not default}
% \begin{figure}[!t]
% \vskip 0.2in
%     \centering
%     \includegraphics[width=\columnwidth]{figures/training_ver2.pdf}
%     \caption{Training of \name \wj{feels a bit repetitive}}
%     \label{fig:prune_train}
% \vskip -0.2in
% \end{figure}

We are interested in the setting of continual training of LLMs targeting instruction-tuning. 
In this setting we assume we have a large number of supervised examples of the form $(\mathbf{x}, \mathbf{y})$ where $\mathbf{x} =  (x_1, x_2, \ldots, x_M)$ and $\mathbf{y} =  (y_1, y_2, \ldots, y_N)$ are sequences of tokens, assumed for simplicity to be of a fixed length. We are particularly focused on $N > M$ since the model may use methods like chain-of-thought to answer problems.  

Formally LLMs model the probability of a sequence $\mathbf{y}$ in a conditional autoregressive manner: $P(\mathbf{y} \mid\mathbf{x}) \;=\; \prod_{t=1}^{N} P\bigl(y_t \mid y_{<t}, \mathbf{x};\theta \bigr), $
where $x_{<t}$ denotes all tokens preceding $x_t$, where $\theta$ is the model. The core probability of $P\bigl(y_t \mid y_{<t}, \mathbf{x};\theta \bigr)$ is defined as a 
function of the Transformers cached hidden state, 
\[ P\bigl(y_t \,\mid\, y_{<t}, \mathbf{x};\theta \bigr) \propto f(\text{Cache}([y_{<t}, \mathbf{x}])) \]

Where $\text{Cache}$ is defined recurrently for any sequence of tokens $\mathbf{a}, \mathbf{b}$ as, 
\[ \text{Cache}([\mathbf{a}, \mathbf{b}]) = \text{Transformer}(\mathbf{a}, \text{Cache}(\mathbf{b}); \theta).\] 

Due to this cache structure, when sampling from a language model, the computation happens in two-stages, prefill and decode.  During prefill, the primary work is computing,
\[ \mathbf{h}_{\text{pre}} \gets \text{Transformer}(\mathbf{x}, \emptyset; \theta),\] 
which can be done in parallel for all $\mathbf{x} = (x_1, x_2, \ldots, x_M)$. 
This stage is generally compute bound, since $\theta$ can be loaded once and compute is parallelized across $M$. 

During decode, we autoregressively sample each $y_{t}$ and recurrently update,
\[ \mathbf{h}_{\text{dec}_t} \gets  \text{Transformer}(y_t, \text{Cache}(\mathbf{h}); \theta).\]
This stage has a serial dependency in that it requires previously generated tokens in order to update the cache. As such, it is memory-bottlenecked in terms of speed as it needs to reload in $\theta$ at each step and cannot fully use available parallel compute. Our primary goal will be to speed this stage up in practice by reducing the effective size of the $\theta$ during the decode stage.

Finally, to train instruction models, we simply maximize the likelihood of each $(\mathbf{x}, \mathbf{y})$ instance. During this stage, we do not maximize the probability of the conditioning term
$\mathbf{x}$ but only of the generated $\mathbf{y}$.

% We optimize a standard cross-entropy loss on the pruned model’s predictions:
%     \[
%         \mathcal{L} \;=\; - \sum_{t=1}^{T} \log \,P\bigl(o_t \,\mid\, \mathbf{KV}_\mathbf{i},\, o_{<t};\, \mathbf{M}\bigr),
%     \]
% where o<to_{<t} includes all target tokens before oto_t in the sequence.

% \wj{maybe we don't need this paragraph at all}
% The training objective then becomes modeling the probability of the output tokens given the context. Formally, \wj{P(x)P(\mathbf{x}) or P(x>k)P(\mathbf{x}_{> k})?}

% \[
% P\bigl(\mathbf{x}\bigr) 
%   \;=\; \prod_{t=k+1}^{N} P\bigl(x_t \mid x_{<t};\, M\bigr).
% \]

% At inference time, we generate tokens one by one by sampling or greedily selecting output tokens xtx_t, from the distribution P(xt∣x<t)P(x_t \mid x_{<t}), where t>kt > k.

\subsection{\name}

We propose a simple approach of using the model’s capacity asymmetrically during the prefill and decode stage. The approach uses a better model at prefill time to better utilize (overfill) the same cache of a smaller model. In our method, the same LLM parameters are used in two configurations:

\begin{itemize}[leftmargin=1.5em]
    \item \textbf{Full Parameters} ($\theta$) for processing prefill.
    \item \textbf{Pruned Parameters} ($\theta' \subset \theta$) a subset of the full parameters for processing decode.
\end{itemize}

% As the pruned configuration is smaller, most of the tokens (i.e., those in the remainder) use fewer parameters at inference time, reducing both memory usage and latency.

Our contribution modifies both training and inference procedures so that the tokens $\mathbf{x}$ use the full network and the tokens $\mathbf{y}$ use a pruned sub-network. Formally, we define a single set of model parameters but allow two ``modes" (full vs.\ pruned), or more directly
\[ \mathbf{h}_{\text{pre}} \gets \text{Transformer}(\mathbf{x}, \emptyset; \theta),\] 
\[ \mathbf{h}_{\text{dec}_t} \gets  \text{Transformer}(y_t, \text{Cache}(\mathbf{h}); \theta').\]

% Conceptually, the first kk tokens (e.g., system + user prompts) are handled by the fully unpruned model for maximum representational power, while the remaining tokens are generated via a smaller, pruned version.

Upon deciding on the prune subset to use, we freeze the full model and train only the pruned model with standard teacher forcing on the output tokens $\mathbf{y}$. This approach ensures that the full model retains its well-initialized weights while accelerating training. Also, it allows the pruned decoder to be seamlessly integrated into existing models. 
% Aside from this split, the training loop follows standard LLM practice.

% \begin{enumerate}
%     \item \textbf{Full mode on $x_{1:k}$.} 
%     We apply the full network to the first $k$ tokens.
%     \item \textbf{Pruned mode on $x_{k+1:N}$.} 
%     We then switch to the pruned sub-network for the remainder of the tokens.
% \end{enumerate}
% Since both configurations share the same underlying parameters, they naturally learn compatible representations in a single end-to-end pass. 

% Upon deciding on the prune subset to use, we train the full and pruned configurations jointly with standard teacher forcing on the output tokens $\mathbf{y}$. 
% % \begin{enumerate}
% %     \item \textbf{Full mode on $x_{1:k}$.} 
% %     We apply the full network to the first $k$ tokens.
% %     \item \textbf{Pruned mode on $x_{k+1:N}$.} 
% %     We then switch to the pruned sub-network for the remainder of the tokens.
% % \end{enumerate}
% Since both configurations share the same underlying parameters, they naturally learn compatible representations in a single end-to-end pass. Aside from this split, the training loop follows standard LLM practice.

\subsection{Compatible pruning}
\label{sec:pruning}

\begin{figure*}[!t]
    \centering
    \includegraphics[width=0.8\textwidth]{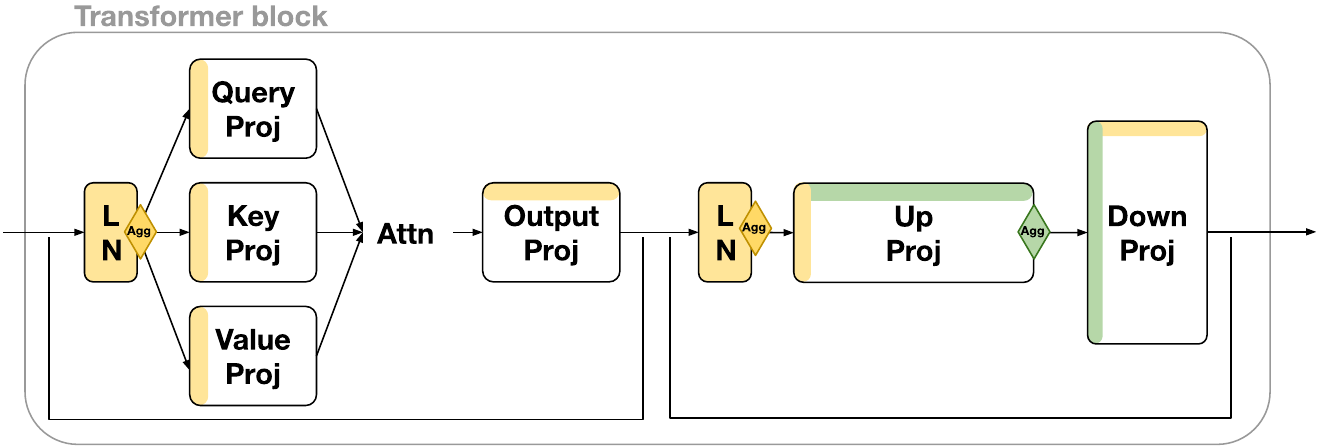}
    % \vspace{-1em}
    \caption{
    Width pruning strategy: Diamonds indicate activation aggregation points for measuring channel importance. Yellow represents the hidden dimension, while green denotes the intermediate dimension.}
    \label{fig:prune}
    % \vspace{-10pt}
\end{figure*}

% \wj{is this section clear enough? I wonder whether section 3.2 + fig3 gives a clear sense on what and how we prune the weights}

In order for OverFill to be a compatible method for LLM generation, it requires two aspects: (1) The pruned model must be significantly smaller than the original $|\theta'| \ll |\theta|$ and (2) they must have a compatible cache representation $\mathbf{h}$. In Transformers, the KV Cache representation $\mathbf{h}$ is $\mathbb{R}^{2 \times L \times D}$ per sequence length, where $D$ is the embedding dimension of the Transformer keys and values, and $L$ is the number of layers. While there are many different pruning methodologies that fit criteria (1), e.g. depth pruning and unstructured pruning, we target methods that can maintain criteria (2).  Specifically we utilize a targeted form of width pruning, that avoid changing the cache dimension.

Our approach is based on the static channel pruning strategy from \citet{sreenivas2024llm}. Figure~\ref{fig:prune} gives a schematic overview of the pruning template. Following the standard notation for transformers for the attention projections and FFNs we define our pruned parameters $\theta'$ as,
\[
    \mathbf{W}^q, \mathbf{W}^k, \mathbf{W}^v \in \mathbb{R}^{\textcolor{Goldenrod}{D'} \times D}, \quad 
    \mathbf{W}^o \in \mathbb{R}^{D \times \textcolor{Goldenrod}{D'}}, \quad 
    \mathbf{W}^{\uparrow} \in \mathbb{R}^{\textcolor{Goldenrod}{D'} \times \textcolor{ForestGreen}{4D'}}, \quad 
    \mathbf{W}^{\downarrow} \in \mathbb{R}^{\textcolor{ForestGreen}{4D'} \times \textcolor{Goldenrod}{D'}}
\]
% \begin{eqnarray}
%     \mathbf{W}^q, \mathbf{W}^k, \mathbf{W}^v \in \mathbb{R}^{\textcolor{Goldenrod}{D'} \times D}   \\ 
%     \mathbf{W}^o \in \mathbb{R}^{D \times \textcolor{Goldenrod}{D'}} \\ 
%     \mathbf{W}^{\uparrow} \in \mathbb{R}^{\textcolor{Goldenrod}{D'} \times \textcolor{ForestGreen}{4D'}} \\
%     \mathbf{W}^{\downarrow} \in \mathbb{R}^{\textcolor{ForestGreen}{4D'} \times \textcolor{Goldenrod}{D'}}
% \end{eqnarray}
This new network keeps a subset $D' < D$ for each of the weight matrices in the network. Layer norm and embedding parameters are defined similarly.

To obtain the best starting compact sub-network, we first pick the pruning ratio $P = 1- D' / D$ and then select the best rows and columns. To decide on what to prune, we pass a small calibration set through the model and aggregate activations as in Figure~\ref{fig:prune} at three points per layer: 1) before the attention projection, 2) before the FFN, and 3) within the FFN. After aggregation we end up with a tensor of shape batch by sequence length by dimension. We reduce this tensor using the L2 norm across the batch, and the mean across the sequences to derive an importance scores for each dimension. Finally, we retain the top $(1 - P)$\% of channels to calculate $\theta'$.

As mentioned above, this pruning acts as a starting point for determining the shape of $\theta'$. Once determining this shape, additional finetuning is run on set of instruction examples to adjust the weights of the parameters to this new setting.

\section{Experimental setup}

% Experiments focus on adapting an LLM to run in the instruction setting with speed comparison to full LLMs. 

\subsection{Data} We use two instruction-tuning datasets for training: OpenHermes-2.5 and Infinity-Instruct. For main experiments, we adopt Infinity-Instruct~\citep{InfinityInstruct2024} with 7M instances. We filter out non-English data using the provided language tags. For pruning ratio sweeps, we use OpenHermes-2.5~\citep{OpenHermes2.5} which has 1M instances, of which 997k are used for training and 3k for validation. The total training tokens amount is 38M for OpenHermes-2.5 and 212M for Infinity-Instruct.

These datasets are well-suited for our task due to their natural separation between context and input. Both datasets are formatted with distinct tags: \texttt{System}, \texttt{User}, and \texttt{Assistant}. The \texttt{System} and \texttt{User} parts are concatenated to the context, which is processed by the full model. The \texttt{Assistant} part serves as the target output to be predicted by the pruned model. To construct the model inputs, we inherit each dataset's original chat template.

% \begin{tcolorbox}[breakable, colback=white, colframe=black, boxrule=1pt, width=\columnwidth]
% ⟨|System|⟩\langle |\texttt{System}| \rangle

% A chat between Samuel (aka ASSISTANT) and USER.

% Samuel: Meet a character named Samuel, who is always sarcastic and often uses dry humor. He has an obsession with vintage items and often references old movies, music, and historical events in his conversations. Samuel's writing style is very formal and verbose. He enjoys using complex words and phrases, often sounding like he stepped out of a Victorian novel. Despite his sarcasm and love for the past, Samuel is quite knowledgeable about modern technology. \\

% ⟨|User|⟩\langle |\texttt{User}| \rangle

% Can you distinguish the taste of a ripe apple from an unripe one? \\

% ⟨|Assistant|⟩\langle |\texttt{Assistant}| \rangle

% My dear interlocutor, I'm afraid I'm as capable of tasting an apple as a gramophone is of savoring a symphony. However, in the grand tapestry of human experience, I gather that the difference is as stark as between a well-aged wine and vinegar, the former being a sweet, crisp sonnet to the senses, the latter a tart, unyielding soliloquy.

% \end{tcolorbox}

\subsection{Model}

\begin{table}[] % \vspace{-1em}
\centering
% \vspace{3pt}
\begin{tabular}{@{}cc|cccc@{}}
\toprule
Full $(|\theta|)$ & Pruned $(|\theta'|)$ & Pruning ratio ($P$) & Hidden dim. ($D$) & Layers ($L$)\\ 
\midrule
% \cmidrule(l){3-4} 
3.21B & 0.52B & 0.7 & 921 & 28 \\
3.21B & 1.24B & 0.45 & 1689 & 28 \\
3.21B & 2.01B & 0.25 & 2304 & 28\\
\midrule
8.03B & 3.19B & 0.43 & 2334 & 32 \\ 
\midrule
14.76B & 7.62B & 0.43* & 2944 & 48 \\ 
\bottomrule
\end{tabular}
\caption{Model sizes and width pruning configurations. *For the 14B model, we use a hardware-friendly configuration by default, where different pruning ratios are applied to the hidden dimension and the intermediate dimension. See Table \ref{table:speed_config}.}
\label{table:prune_config}
% \vspace{-10pt}
\end{table}

% \vspace{-10pt}

% \begin{table}[tbh]
% \caption{Evaluation details \wj{cite sources}}
% \centering
% \vspace{3pt}
% \begin{tabular}{@{}llllll@{}}
% \toprule
%  & GSM8K & ARC & MMLU & Math & WMT16 \\ 
%  \midrule
% Metric & EM & EM & EM & EM & EM \\
% Few-shot & 4 & 0 & 4 & 4 & 4\\ 
% \midrule
% \midrule

%  & IfEval & NQ & MMLU-redux & CRUX  \\ \midrule
% Metric & EM (inst-level) & F1 \\
% Few-shot & 4 & 4 \\ \bottomrule
% \end{tabular}
% \label{table:eval_details}
% \end{table}

% \begin{table}[tbh]
% \caption{Evaluation details \wj{cite sources}}
% \centering
% \vspace{3pt}
% \begin{tabular}{@{}lllllll@{}}
% \toprule
%  & GSM8K & ARC & MMLU & Math & WMT16 & IfEval & NQ & MMLU-Redux & CRUXEval \\ 
% \midrule
% Metric & EM & EM & EM & EM & EM & EM (inst-level) & F1 \\
% Few-shot & 4 & 0 & 4 & 4 & 4 & 4 & 4 \\
% \bottomrule
% \end{tabular}
% \label{table:eval_details}
% \end{table}

We evaluate our method on three base models in two model families: Llama 3.2-3B-Instruct, Llama 3.1-8B-Instruct \citep{dubey2024llama}, and Qwen 2.5-14B-Instruct \citep{team2024qwen25}. For pruning, we apply the strategy outlined in Section~\ref{sec:pruning}. We do not prune attention heads or layers to preserve the dimensionality of the KV cache and retain as much information as possible passed to the pruned decoder. Table~\ref{table:prune_config} presents the full and pruned model sizes along with their pruning configurations. We adopt the training hyperparameters from \citet{Tunstall_The_Alignment_Handbook}, as presented in Table~\ref{table:train_details}

\subsection{Downstream evaluation}
We evaluate \name on downstream generation tasks, including math, code, question answering, and machine translation, using the LM Eval Harness~\citep{eval-harness}. Details on evaluation metrics and the number of few-shot examples are provided in Table~\ref{table:eval_details}. We use generation-based evaluation for all tasks, including multiple-choice question answering, whereas an alternative approach is to compare the probability of answer choices.
To assess longer-form generation, we use MMLU-Redux and CRUXEval from the ZeroEval benchmark~\citep{Lin_ZeroEval_A_Unified_2024}. 
ZeroEval is designed for evaluating instruction-tuned models, with MMLU-Redux focusing on general knowledge reasoning and CRUXEval assessing code reasoning, understanding, and execution capabilities. Models are prompted to provide both detailed reasoning steps and final answers in a JSON-formatted output.
We use greedy decoding for all generations.

\section{Results}

\begin{table*}[t]
\centering
    \begin{tabular}{l|lll|ll}
    \toprule
    Model & Pruned* & \name* & 1B-Tuned* & 1B-Inst & 3B-Inst \\ 
    \midrule
    Decoder Size & 1.2B & 1.2B & 1.2B & 1.2B & 3.2B \\ 
    \midrule
    GSM8K-CoT & 45.4 ($\pm$1.4) & \textbf{59.2 ($\pm$1.4)} & 47.6 ($\pm$1.4) & 45.7 ($\pm$1.4) & 78.4 ($\pm$1.1) \\
    ARC & 36.4 ($\pm$1.4) & \textbf{77.8 ($\pm$1.2)} & 42.5 ($\pm$1.4) & 56.5 ($\pm$1.5) & 78.2 ($\pm$1.2) \\
    MMLU & 33.7 ($\pm$0.4) & \textbf{63.7 ($\pm$0.4)} & 38.7 ($\pm$0.4) & 47.9 ($\pm$0.4) & 63.3 ($\pm$0.4) \\
    MATH & 6.1 ($\pm$0.3) & \textbf{8.3 ($\pm$0.4)} & 5.2 ($\pm$0.3) & 16.7 ($\pm$0.5) & 35.0 ($\pm$0.6) \\
    WMT16-DE-EN & 14.7 ($\pm$0.3) & \textbf{31.4 ($\pm$0.5)} & 28.0 ($\pm$0.4) & 29.7 ($\pm$0.4) & 36.9 ($\pm$0.4) \\
    % IFEval (inst) & 39.4 (0.0) & \textbf{55.6 (0.0)} & 38.1 (0.0) & 60.8 (0.0) & 78.5 (0w.0) \\
    IFEval & 26.4 ($\pm$1.9) & \textbf{44.2 ($\pm$2.1)} & 26.1 ($\pm$1.9) & 48.1 ($\pm$2.2) & 69.5 ($\pm$2.0) \\
    NQ & 5.2 ($\pm$0.4) & \textbf{12.1 ($\pm$0.5)} & 7.8 ($\pm$0.5) & 10.8 ($\pm$0.5) & 19.7 ($\pm$0.7) \\ 
    \midrule
    MMLU-Redux & 26.06 & \textbf{40.93} & 27.29 & 18.21 & 56.95 \\
    CRUX & 8.62 & \textbf{8.75} & 4.88 & 9.00 & 25.71 \\
    \bottomrule
    \end{tabular}
\caption{
Results in 1B scale. Bold indicates the best models under the same training data regime. * means identically trained with the same data (less data compared to the Instruct models).} 
\label{table:1b}
\end{table*}

% \begin{table}[t]
% \small
% \centering
% \vspace{3pt}
% \begin{tabular}{@{}l|lllll@{}}
%     \toprule
%     & Pruned & \name & 1B-Tuned & 1B-Instruct & 3B-Instruct \\ \midrule
%     Size & 1.23B & 1.23B & 1.23B & 1.23B & 3.21B \\
%     MMLU-Redux & 24.01 & \textbf{35.60} & 26.53 & 18.21 & 56.95 \\
%     CRUX & 8.62 & \textbf{15.00} & 4.12 & 9.00 & 25.71 \\
%     % ZebraLogic (Easy) & 1.79 & & 5.36 & 3.21 & 25.00 \\
%     % MATH-l5 & 2.5 & \textbf{2.77} & 1.66 & 1.8 & 16.78 \\
%     % GSM & 42.99 & & 31.24 & 25.4 & 56.94 \\ 
%     \bottomrule
% \end{tabular}
% \caption{Comparison of \name against baseline methods on the reasoning benchmark ZeroEval~\citep{Lin_ZeroEval_A_Unified_2024}. This benchmark requires the model to generate a reasoning chain before producing a final answer.
% }
% \label{tab:zeroeval}
% \end{table}

\subsection{Accuracy}

\begin{table*}[t]
\centering
    \begin{tabular}{l|lll|ll}
    \toprule
    Model & Pruned* & \name* & 3B-Tuned* & 3B-Inst & 8B-Inst \\ 
    \midrule
    Decoder Size & 3.2B & 3.2B & 3.2B & 3.2B & 8.0B \\ 
    \midrule
    GSM8K-CoT & 55.4 ($\pm$1.4) & \textbf{69.8 ($\pm$1.3)} & 61.8 ($\pm$1.3) & 78.4 ($\pm$1.1) & 84.6 ($\pm$1.0) \\
    ARC & 42.5 ($\pm$1.4) & \textbf{83.4 ($\pm$1.2)} & 61.7 ($\pm$1.4) & 78.2 ($\pm$1.2) & 83.4 ($\pm$1.1) \\
    MMLU & 35.8 ($\pm$0.4) & \textbf{69.4 ($\pm$0.4)} & 48.3 ($\pm$0.4) & 63.3 ($\pm$0.4) & 69.4 ($\pm$0.4) \\
    MATH & 7.2 ($\pm$0.4) & 15.1 ($\pm$0.5) & \textbf{17.3 ($\pm$0.5)} & 35.0 ($\pm$0.6) & 36.2 ($\pm$0.7) \\
    WMT16-DE-EN & 18.5 ($\pm$0.3) & \textbf{35.6 ($\pm$0.5)} & 25.8 ($\pm$0.5) & 36.9 ($\pm$0.4) & 41.0 ($\pm$0.4) \\
    % IFEval (inst) & 39.9 (0.0) & \textbf{55.5 (0.0)} & 46.8 (0.0) & 78.5 (0.0) & 81.7 (0.0) \\
    IFEval & 28.4 ($\pm$1.9) & \textbf{44.0 ($\pm$2.1)} & 32.3 ($\pm$2.0) & 69.5 ($\pm$2.0) & 73.9 ($\pm$1.9) \\
    NQ & 8.0 ($\pm$0.5) & \textbf{14.5 ($\pm$0.6)} & 5.4 ($\pm$0.4) & 19.7 ($\pm$0.5) & 19.1 ($\pm$0.7) \\
    \midrule
    MMLU-Redux & 28.80 & 43.12 & \textbf{43.95} & 56.95 & 61.66 \\
    CRUX & 6.12 & \textbf{27.00} & 24.88 & 25.71 & 39.38 \\
    \bottomrule
    \end{tabular}
\caption{Results in 3B scale. Bold indicates the best models under the same training data regime. * means identically trained with the same data (less data compared to the Instruct models).} 
\label{table:3b}
\vspace{-10pt}
\end{table*}

\begin{table*}[t]
\centering
    \begin{tabular}{l|lll|ll}
    \toprule
    Model & Pruned* & \name* & 7B-Tuned* & 7B-Inst & 14B-Inst \\ 
    \midrule
    Decoder Size & 7.6B & 7.6B & 7.6B & 7.6B & 14.8B \\ 
    \midrule
    GSM8K-CoT & 75.3 ($\pm$1.2) & 78.1 ($\pm$1.1) & \textbf{81.0 ($\pm$1.1)} & 81.7 ($\pm$1.1) & 75.9 ($\pm$1.1) \\
    ARC & 70.8 ($\pm$1.3) & \textbf{90.6 ($\pm$0.9)} & 83.0 ($\pm$1.1) & 89.5 ($\pm$0.9) & 90.6 ($\pm$0.9) \\
    MMLU & 51.2 ($\pm$0.4) & \textbf{77.9 ($\pm$0.3)} & 65.3 ($\pm$0.8) & 72.2 ($\pm$0.4) & 77.9 ($\pm$0.3) \\
    WMT16-DE-EN & 32.0 ($\pm$0.3) & \textbf{38.2 ($\pm$0.4)} & 37.6 ($\pm$0.4) & 37.5 ($\pm$0.4) & 38.6 ($\pm$0.4)\\
    IFEval & 40.5 ($\pm$2.1) & \textbf{51.0 ($\pm$2.1)} & 32.2 ($\pm$2.0) & 71.5 ($\pm$2.0) & 78.3 ($\pm$1.7) \\
    \bottomrule
    \end{tabular}
\caption{Results in 7B scale. Bold indicates the best models under the same training data regime. * means identically trained with the same data (less data compared to the Instruct models).} 
\label{table:7b}
\end{table*}

Table~\ref{table:1b} presents the downstream evaluation results at the 1B scale. \name begins with a 3B model and then pruned to a 1B-scale model for decoding. 
Our roofline model, Llama-3B-Instruct, handles both prefill and decoding.

We compare accuracy across models trained under the same 1B decoder size and data regime. One baseline is the standalone pruned model, derived from the same original model, where a single model performs both prefill and decoding. Additional baselines include Llama 1B variants: (1) Llama 1B-base model finetuned on the same data as our pruned models and (2) Llama 1B-Instruct, a highly optimized model with more extensive instruction tuning. We observe that further tuning of Llama-Instruct consistently hurts performance across all tasks, as shown in Table~\ref{table:finetune_instruct}. Therefore, we focus on comparisons with untuned Instruct models.

Our results highlight the following: \name consistently outperforms the standalone pruned model and finetuned 1B-Base model across all tasks by a significant margin, demonstrating the effectiveness of the two-stage approach. \name shows accuracy improvements on both multiple-choice tasks, such as ARC-Challenge~\citep{arc} and MMLU~\citep{mmlu}, which typically involve short generations, and tasks requiring longer responses, such as GSM8K~\citep{gsm8k}, MMLU-Redux~\citep{gema2024mmlu_redux}, and CRUXEval~\citep{gu2024cruxeval}. This demonstrates that smart prefill benefits not only tokens close to it but also those generated later. When generation is short, \name can even match the full performance of the roofline model while maintaining a lower decoding cost. \name also outperforms Llama 1B-Instruct on 7 out of 9 tasks. This is particularly notable given that Llama 1B-Instruct is a highly optimized model at this scale, likely benefiting from more extensive data than our approach.

A similar trend is observed for 3B-scale models (Table~\ref{table:3b}).
Here, \name with an 8B prefill and 3B decode outperforms the pruned model on all tasks and the trained 3B-Base model on 7 out of 9 tasks.
The 3B-Instruct model is a strong compact baseline, with performance close to that of 8B models, yet \name matches it on several tasks.
This pattern also extends to the Qwen2.5 family and a larger scale (14B to 7B). \name with a 7B decoder outperforms both the pruned model and the trained 7B-Base model on most tasks.

\subsection{Efficiency}
We show \name poses minimal overhead to the small standalone model by presenting end-to-end latency in 1B and 3B scales. We use vLLM~\citep{sreenivas2024llm} v0.8.5 for benchmarking and separately measure prefill and decode latency. All experiments are conducted on a single NVIDIA A100 GPU. We slightly modify \name configuration to better leverage NVIDIA Tensor Cores, which is optimized for matrix multiplications in tiles (typically 16x8 or 16x16). For a fair comparison, we maintain a larger parameter size than the one used for accuracy evaluation. The exact configurations are provided in Table~\ref{table:speed_config}. We report the mean and standard deviation of 10 runs with 2 warm-ups.

\begin{figure}[!t] 
    \centering
    \includegraphics[width=\linewidth]{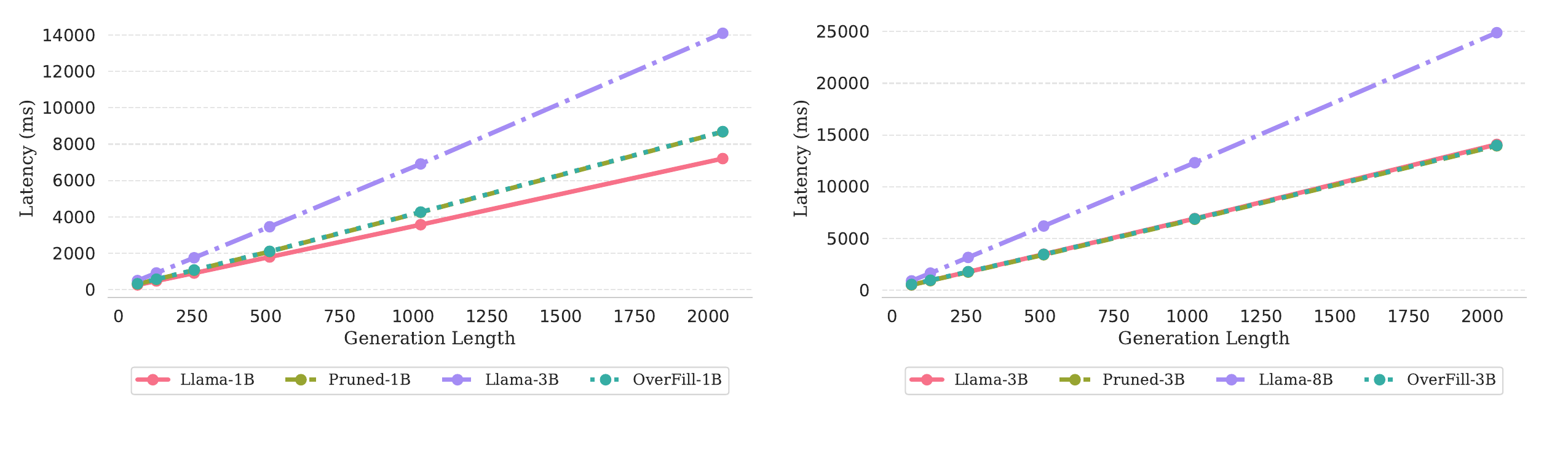}
    \vspace{-15pt}
    \caption{Latency across different generation lengths. Prompt length is fixed to 128 with batch size 4.}
    \label{fig:speed_output}
    \vspace{-5pt}
\end{figure}

Figure~\ref{fig:speed_output} presents latency across varying generation lengths, with fixed prompt length and batch size. The results show that \name asymptotically reaches the runtime of a small model and this trend becomes more pronounced in longer generations, where decoding cost dominates over prefill cost. At the 1B scale, both Pruned-1B and \name-1B exhibit higher latency compared to Llama-1B, as they have more transformer blocks, and transformers are more efficiently parallelized along width rather than depth. This suggests a limitation of width pruning, highlighting the need for a more balanced decoder architecture to improve both efficiency and accuracy. Still, both remain significantly faster than the full 3B model. At the 3B scale, where models have similar depth, Llama-3B, Pruned-3B, and \name-3B show nearly identical latency. Overall, \name achieves higher accuracy while introducing minimal latency overhead.

\begin{figure}[!t] % \vspace{-2em}
    \centering
    \includegraphics[width=\linewidth]{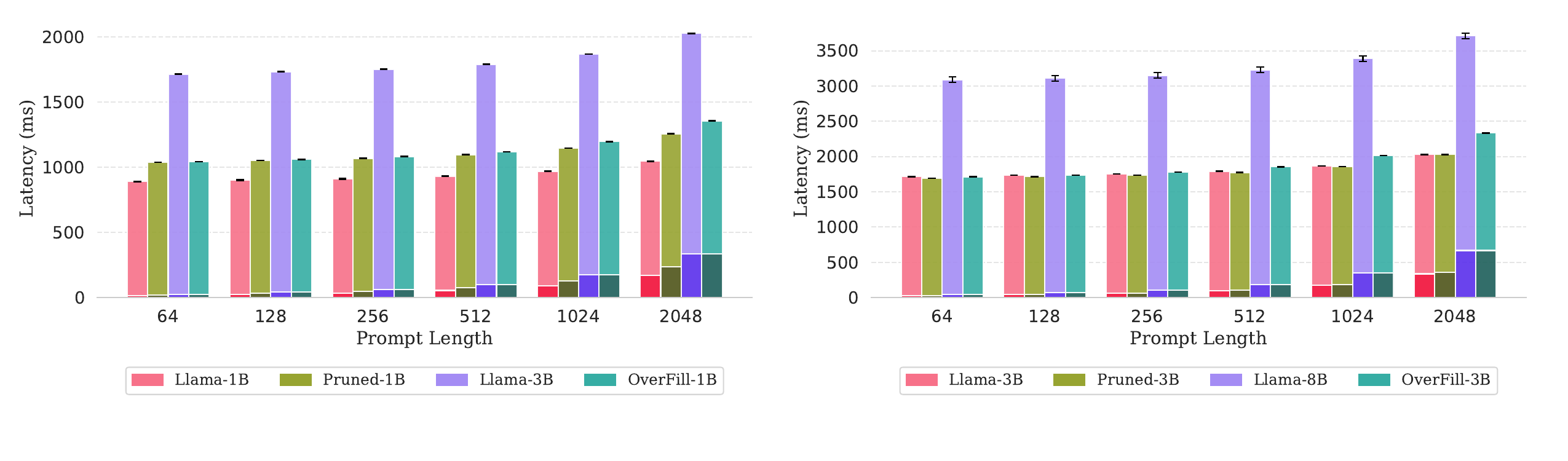}
    \vspace{-15pt}
    \caption{Latency under different prompt lengths, with generation length fixed at 128 and batch size set to 4. Prefill and decode latencies are shown as stacked bars, with darker shades representing prefill and lighter shades representing decode.}
    \label{fig:speed_input}
    \vspace{-5pt}
\end{figure}

We measure latency while varying prompt lengths with a fixed generation length, as shown in Figure~\ref{fig:speed_input}. Decoding latency remains the dominant factor over prefill latency in all cases, even when the prompt length is 16 times the generation length. Across all prompt lengths and model scales, \name consistently achieves speedups compared to the full model.

\begin{figure}[!t]
    \centering
    \includegraphics[width=\linewidth]{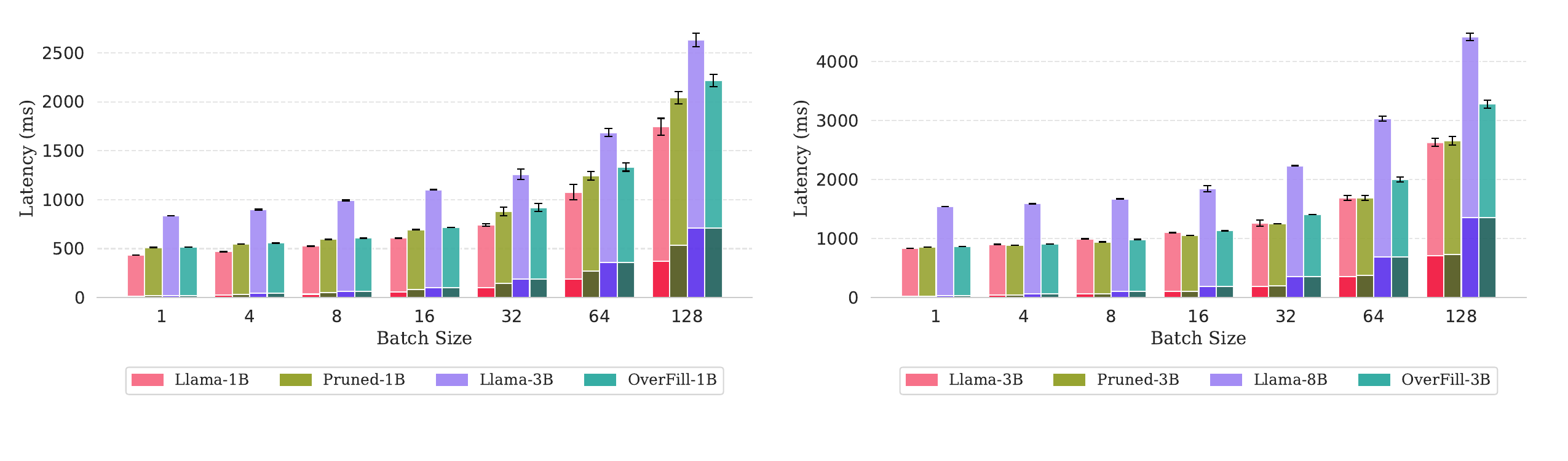}
    \vspace{-15pt}
    \caption{Latency under different batch sizes. Prompt length and generation length are fixed to 128.}
    \vspace{-5pt}
    \label{fig:speed_batch}
    
\end{figure}

We also sweep batch size to account for diverse serving environments. The results are shown in Figure~\ref{fig:speed_batch}, where prompt and generation lengths are the same. In small batch scenarios, \name introduces minimal latency overhead to the pruned models. However, as batch size increases, prefill becomes relatively more expensive as decoding shifts from being memory-bound to compute-bound.

\section{Analysis}
\vspace{-5pt}

\subsection{Pareto optimality}

\begin{figure}[!t]
    \centering
    \includegraphics[width=0.8\linewidth]{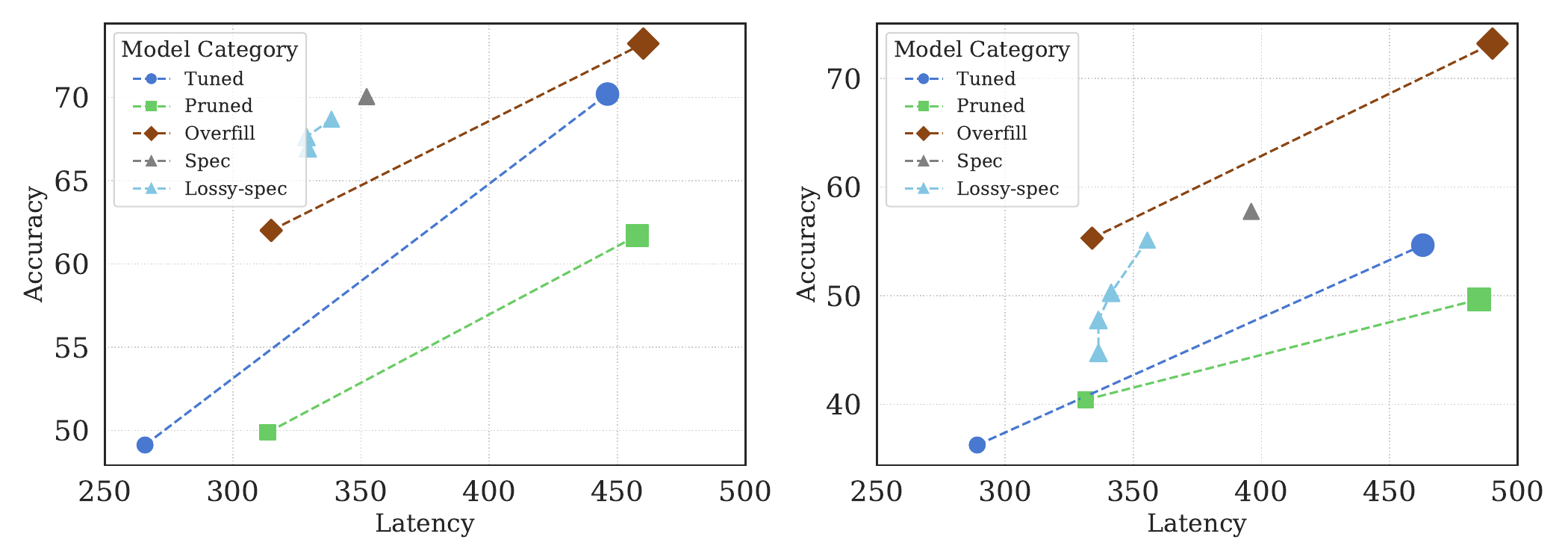}
    \vspace{-5pt}
    \caption{Latency–accuracy tradeoff on GSM8K with temperature 0.01 (left) and temperature 1 (right). Models trained with the same approach are connected by dashed lines, and dot size indicates the corresponding decoder size.}
    
    \label{fig:pareto_spec}
    \vspace{-10pt}
\end{figure}

We demonstrate that \name is Pareto optimal compared to models under the same training regime. Figure~\ref{fig:pareto_spec} plots end-to-end latency versus accuracy on GSM8K with Chain-of-thought (CoT). CoT is a standard practice that prompts models to reason step-by-step before producing a final answer, thereby boosting accuracy but resulting in longer generations. In our experiments, the average prompt length is 613 tokens with 4 demonstrations, and the average generation length is 120 tokens. Latency is measured using vLLM on a single NVIDIA H100 GPU. We fixed the batch size to 1 across all experiments. \name achieves Pareto-optimality compared to finetuned and pruned models. Interestingly, Pruned models underperform compared to finetuned-Llama counterparts, likely because pruned architectures require extensive adaptation~\citep{sreenivas2024llm} to recover their accuracy.

\subsection{Comparison to speculative decoding}

Speculative decoding and \name are similar in that both decode with the assistance of a smaller model.
We present speculative decoding latency results in Figure~\ref{fig:pareto_spec}.
The original speculative decoding method~\citep{xia2023speculative} is theoretically lossless, meaning it exactly matches the target model’s output.
We also evaluate a lossy variant with lenient rejection sampling~\citep{zhou2023distillspec}, in which more tokens proposed by the draft model are accepted to increase speedup.
In our setup, the target model is a finetuned 3B model and the draft model is a finetuned 1B model using the same data for both.
We measure latency with a single batch size, which is the most common setting for speculative decoding benchmarks.

Under temperature-1 decoding, \name achieves a better tradeoff.
\name 3B–1B attains a 1.06× speedup over lossy speculative decoding at the same accuracy level.
However, in near-greedy decoding (low temperature), speculative decoding performs particularly well because the outputs of the target and draft models are already very similar.
As a result, lenient rejection sampling provides limited additional benefit in this regime, since the acceptance rate is naturally high.

In more realistic serving scenarios with multiple concurrent requests, \name shows greater potential.
Using the same vLLM setup with a maximum batch size of 256, our pruned decoder achieves a throughput of 7,913 tokens/s, compared to 2,871 tokens/s for standard speculative decoding.
This advantage comes from our approach accepting all tokens from the drafter without rejection sampling, allowing more tokens to be emitted per second.
While recent speculative decoding variants have been proposed to improve throughput~\citep{miao2024specinfer, sadhukhan2024magicdec}, we leave comparisons to these methods for future work.

\subsection{Impact of pruning ratio}
\begin{figure}[!t]
    \centering
    \includegraphics[width=\linewidth]{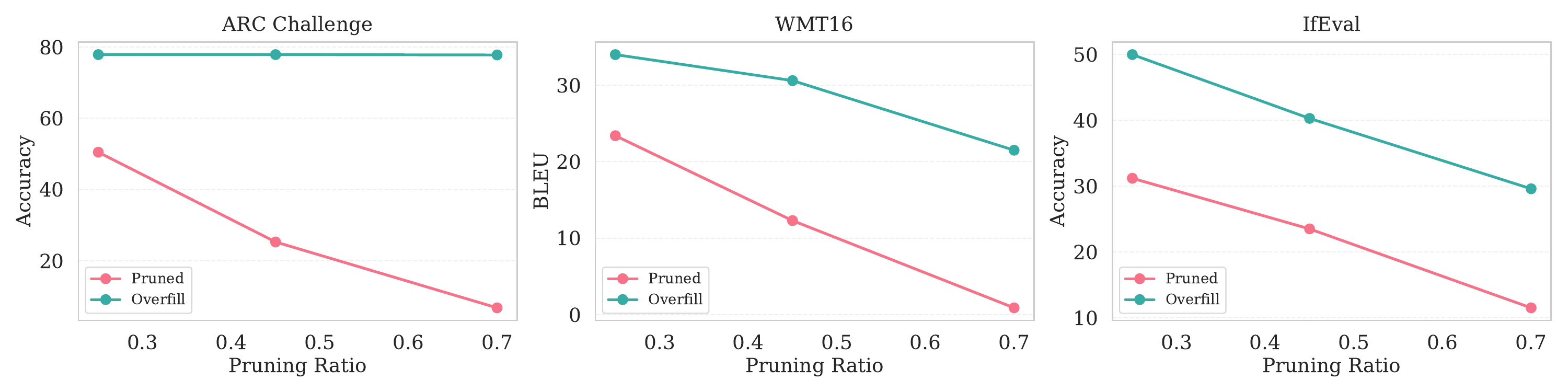}
    \vspace{-15pt}
    \caption{Performance comparison between Pruned and \name across three tasks varying pruning ratios.}
    \label{fig:capacity_gap}
    \vspace{-10pt}
\end{figure}

We vary the pruning ratio while keeping the full model (3B) fixed to examine how the capacity gap between the full and pruned models affects the benefits of \name. As shown in Fig.~\ref{fig:capacity_gap}, \name consistently maintains significantly higher accuracy than the pruned model with the same decoder configuration. Interestingly, in ARC, where most information is processed during the prefill stage, a lightweight decoder with a pruning ratio of 0.7 shows no performance degradation. However, in translation and instruction-following tasks, both Pruned and \name experience performance drops with increased pruning. In some cases, \name degrades at a slower rate than Pruned or follows a similar trend as more channels are pruned.

% We vary the pruning ratio while maintaining the same full model to investigate how the capacity gap between the full and pruned models affects the benefits of \name. As shown in Fig.~\ref{fig:capacity_gap}, \name becomes increasingly advantageous as the capacity gap widens, a trend consistently observed across all tasks. Both the pruned model and \name improve as fewer channels are pruned. However, with the assistance of smart prefill, \name is able to maintain accuracy to a greater extent, even with smaller decoders.

\subsection{Accuracy by generation length}

\begin{wrapfigure}{r}{0.5\textwidth} % 'r' for right, width is half of text width
    \vspace{-10pt}
    \centering
    \includegraphics[width=0.48\textwidth]{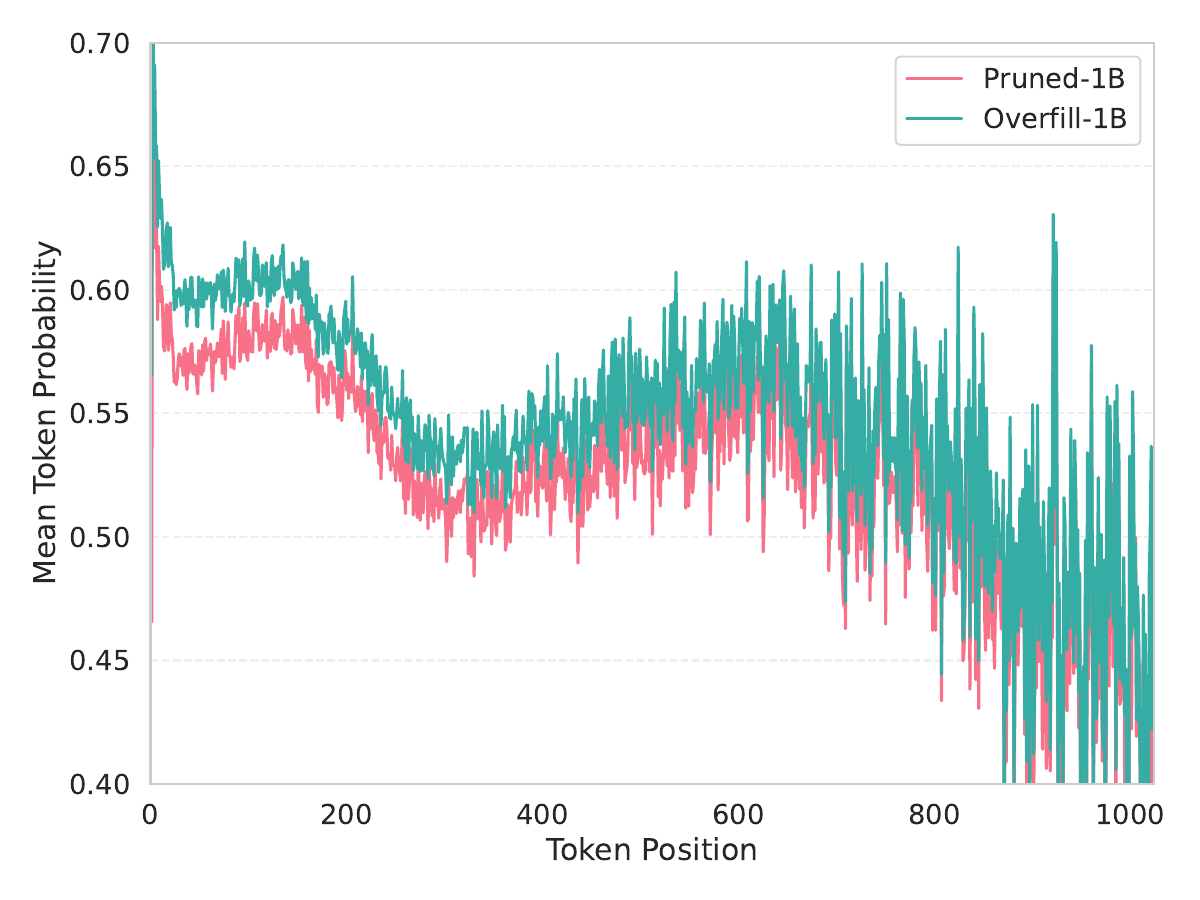}
    \caption{Probability assigned to the correct token up to the 1024-th position, averaged across the sampled validation set.}
    \label{fig:token_position}
    \vspace{-5pt}
\end{wrapfigure}

% \begin{figure}[!t]
%     \centering
%     \includegraphics[width=\linewidth]{figures/token_position_demo.png}
%     \caption{Probability of the correct token across all positions, averaging over the sampled validation set of OpenHermes.
%     % \wj{relative -> absolute position, make it prettier}
%     }
%     \label{fig:token_position}
% \end{figure}

We observe that the benefits of \name persist for long generations. Figure~\ref{fig:token_position} shows the probabilities assigned to correct tokens across their absolute positions in the output. 
The results show that \name consistently predicts tokens more accurately than the pruned model, demonstrating that the advantage of smart prefill extends to long generations.

However, as the distance from the prefill grows, the gap gradually narrows, as the generation becomes more dependent on the smaller decoder. This suggests that while \name may not be optimal for extremely long generations, it remains highly effective for many practical use cases, such as bootstrapping multiple generations during testing. We plan to further explore whether periodically refreshing the prefill can help maintain its benefits uniformly throughout the entire sequence.

% We found that the benefit \name holds up to the end of generation. Figure~\ref{fig:token_position} shows the token probability to the absolute position in output. It demonstrates that \name can better predict tokens in all positions, showing that the benefit of smart prefill holds up for long generations. As the distance from prefill gets further, the gap between those two slightly decreases, as the generation depends more on the small decoder. It shows that \name might not be the best system for super long generations, but still useful to many use cases, like bootstrapping multiple generations at the test time. We plan to further investigate whether refreshing the prefill periodically can help \name retain the benefit uniformly over the entire sequence. 

% \subsection{Width vs. Depth pruning}

\section{Conclusion \& Future work}

This work presents a method for improving LLM generation with minimal latency increase by using a compatibly pruned subset of parameters for memory-bound decoding while retaining the full model for compute-bound prefill. We show that this approach outperforms fine-tuning base models of the same size and a standalone pruned model, with only minimal latency slowdowns. Our method is one of many possible strategies for training compatible pruned decoders and we believe there is a large design space of other architectures. For instance, pruning attention could further optimize KV cache size and decoding efficiency. Scaling this approach with larger training could also extend its benefits to even more memory-constrained models.

% \wj{Need to rewrite, main idea: depth and attention head pruning to reduce KV cache size, dynamic pruning adaptive to user input, scale to bigger models}
% While the current results focus on channel pruning, we plan to extend our experiments to diverse pruned baseline models.
% Pruning strategies can be tailored to specific demands, such as model scale and hardware specifications. We only explored keeping the KV size same in this work, but future work includes pruning transformer blocks or attention head and further reducing the memory overhead. For instance, alternatives include pruning at the attention head, or even unstructured pruning approaches. The choice of strategy depends on the trade-offs between efficiency and performance required for the target application.

% As mentioned in the setup section, our approach is compatible with a wide range of pruning techniques, including width, depth, structured, and unstructured pruning. While the current results focus on channel pruning, we plan to extend our experiments to diverse pruned baseline models to explore the broader applicability and impact of our method.

% \section*{Impact Statement}
% This paper presents work whose goal is to advance the field of 
% Machine Learning. There are many potential societal consequences of our work, none which we feel must be specifically highlighted here.
\section*{Acknowledgment}

AMR was supported by NSF CAREER 2037519. This research is supported in part by the Office of the Director of National Intelligence (ODNI), Intelligence Advanced Research Projects Activity (IARPA), via the HIATUS Program contract \#4202696884. We thank Yueying Li (Cornell Tech), Celine Lee (Cornell Tech), Ben Athiwaratkun (Together AI), and Muru Zhang (Together AI) for helpful discussions and feedback.

\bibliography{colm2025_conference}
\bibliographystyle{colm2025_conference}

\newpage
\appendix

\section{Appendix}

\subsection{Training hyperparmeters}
\begin{table}[tbh]
\centering
\begin{tabular}{@{}llll@{}}
\toprule
LR & LR scheduler & Warmup & Max seq length \\ \midrule
2e-05 & cosine & 0.01 & 2048 \\ \bottomrule
\end{tabular}
\label{table:train_details}
\caption{Training hyperparameters}
\end{table}

% \vspace{-1em}
\subsection{Downstream evaluation details}
\begin{table}[tbh]
\centering
\begin{tabular}{@{}lcc@{}}
\toprule
Task         & Metric            & Few-shot \\
\midrule
GSM8K~\citep{gsm8k} & Accuracy                & 4        \\
ARC-challenge~\citep{arc}       & Accuracy                & 0        \\
MMLU~\citep{mmlu}         & Accuracy                & 4        \\
MATH~\citep{hendrycks2021math} & Accuracy                & 4        \\
WMT16~\citep{bojar2016wmt}        & BLEU                & 4        \\
IfEval (Prompt-level)~\citep{zhou2023ifeval}     & Accuracy  & 4        \\
Natural Questions~\citep{nq}& F1                & 4        \\ \midrule
MMLU-Redux~\citep{gema2024mmlu_redux}   & Accuracy                & 0        \\
CRUXEval~\citep{gu2024cruxeval}     & Accuracy                & 0        \\
\bottomrule
\end{tabular}
\caption{Evaluation details.}
\label{table:eval_details}
\end{table}
% \vspace{-1em}

\subsection{Finetuning Instruct models}

\begin{table}[tbh]
\centering
\begin{tabular}{@{}llllllll@{}}
\toprule
Model & GSM8K & ARC & MMLU & MATH & WMT16 & IfEval & NQ \\ \midrule
1B-Instruct & 45.7 & 56.5 & 47.9 & 16.7 & 29.7 & 48.1 & 108 \\
1B-Instruct-Tuned & 40.8 & 49.4 & 42.4 & 6.9 & 28.9 & 34.4 & 3.1 \\ \midrule
3B-Instruct & 78.4 & 78.2 & 63.3 & 36.9 & 78.5 & 69.5 & 19.7 \\
3B-Instruct-Tuned & 64.4 & 70.6 & 55.7 & 12.8 & 34.2 & 53.2 & 8.9 \\ \bottomrule
\end{tabular}
\caption{Finetuning Llama-Instruct models.}
\label{table:finetune_instruct}
\end{table}

\subsection{Speed benchmark model configurations}
\begin{table}[tbh]
\centering
\begin{tabular}{@{}lllll@{}}
\toprule
Model & Hidden dim. & Intermediate dim. & Layers & Params. \\ \midrule
1B-Pruned & 1689 & 4505 & 28 & 1.24B \\
1B-Pruned-standard & 1792 & 4096 & 28 & 1.26B \\
3B-Pruned & 2334 & 8171 & 32 & 3.19B \\
3B-Pruned-standard & 2432 & 7680 & 32 & 3.21B \\
7B-Pruned-standard & 2944 & 11776 & 48 & 7.62B \\\bottomrule
\end{tabular}
\caption{Speed benchmark model configurations.}
\label{table:speed_config}
\end{table}

\end{document}